# Title: A modality-aware benchmark reveals when spiking neural networks benefit edge sensing

Short title: Benchmarking SNNs in edge sensing

## Authors

Xin Du [1,2,†*], Di Yu [2,3,†], Changze Lv [4], Yuqi Zhang [5], Zhuo Chen [2,3], Wentao Tong [2,3], Helin Zheng [1], Weisong Zhang [1], Xiaofan Zhao [2,3], Linshan Jiang [6], Zhanglu Yan [7], Shijie Ji [8], Hui Fang [5], Jianfeng Feng [9,10], Huajin Tang [2,3], Xiaoqing Zheng [4], Gang Pan [2,3*] and Shuiguang Deng [2,3*]

## Affiliations

1 School of Software Technology, Zhejiang University, Hangzhou, Zhejiang, China
2 State Key Lab of Brain-Machine Intelligence, Zhejiang University, Hangzhou, Zhejiang, China
3 College of Computer Science and Technology, Zhejiang University, Zhejiang, China
4 School of Computer Science and Artificial Intelligence, Fudan University, Shanghai, China
5 School of Information Management and Engineering, Shanghai University of Finance and Economics, Shanghai, China
6 Department of Computer Science and Engineering, Southern University of Science and Technology, Shenzhen, China
7 Institute of Data Science, National University of Singapore, Singapore
8 Division of Engineering and Applied Science, California Institute of Technology, USA
9 Institute of Science and Technology for Brain-Inspired Intelligence, Fudan University, Shanghai, China
10 Department of Computer Science, University of Warwick, Coventry, United Kingdom

† These authors contributed equally to this work.
* Corresponding authors: Xin Du (Email: xindu@zju.edu.cn) or Gang Pan (Email: gpan@zju.edu.cn) or Shuiguang Deng (Email: dengsg@zju.edu.cn)

## Abstract

Edge systems increasingly require models that operate under strict energy, latency, and memory constraints across sensing modalities. Spiking neural networks (SNNs) offer bio-inspired alternatives to artificial neural networks (ANNs), yet their advantages remain inconsistent and debatable. Here, we show that SNN effectiveness is not universal but modality-dependent. We construct a benchmark covering five sensing modalities, 24 datasets, 7 physical devices, and 4 neuromorphic simulation platforms, and evaluate spike encodings, neuron models, and network architectures under unified training and deployment protocols. Across 4,662 evaluations, SNNs achieve ANN-comparable performance in most modalities but exhibit a pronounced advantage in wireless workloads. Spectral and feature-space analyses suggest that this advantage arises from alignment between signal structure and the low-pass filtering behavior of spiking dynamics. Deployment analysis further reveals that SNN efficiency is multi-dimensional. To support reproducibility and future research, we release Soul-NeuSim, an open-source benchmark and deployment framework for edge intelligence and neuromorphic co-design.

## Teaser

SNNs benefit edge sensing selectively, depending on modality-specific signal dynamics.

# MAIN TEXT

## INTRODUCTION

Edge intelligence is moving from single-task inference toward heterogeneous sensing systems that must process diverse workloads under tight constraints on energy, latency, and memory. Such systems increasingly support applications ranging from wearable health monitoring and smart environments to autonomous robotics, wireless sensing, and event-driven perception (1-8). In these settings, the central challenge is no longer whether a model can achieve high accuracy in isolation, but whether it can maintain reliable performance across sensing modalities while remaining deployable on resource-constrained hardware.

Artificial neural networks (ANNs) have become the dominant computational paradigm for edge intelligence, but their dense activation patterns, memory traffic, and parameterized operations can impose substantial costs on embedded platforms (9,10). Spiking neural networks (SNNs), inspired by biological neural computation, offer an alternative through event-driven information processing, sparse activation, and temporal state integration (11,12). These properties have motivated growing interest in using SNNs for low-power edge computing. However, despite these conceptual advantages, the practical benefits of SNNs remain inconsistent across applications. In some tasks, SNNs approach or exceed ANN performance with lower energy consumption, whereas in others they incur accuracy loss, latency overhead, or increased memory footprint (13). This inconsistency suggests that SNN effectiveness cannot be understood simply as a property of the network architecture.

A key reason is that SNN behavior is shaped by design dimensions that are largely absent from conventional ANN evaluation. Spike encoding determines how continuous sensor inputs are converted into temporal spike trains. Neuron dynamics determine how information is filtered, accumulated, and forgotten over time. Network topology determines how spatial, temporal, and cross-channel dependencies are represented. These factors interact with the statistical structure of input data and with the constraints of deployment hardware. As a result, evaluating SNNs only on isolated benchmarks or through aggregate metrics such as spike counts or operation counts provides an incomplete view of when SNNs are actually useful for edge sensing.

This problem becomes especially important across sensing modalities. Vision, audio, motion, wireless, and neuromorphic event data differ substantially in sampling rate, sparsity, noise characteristics, temporal continuity, and frequency structure. Vision workloads often contain dense spatial patterns and temporal redundancy. Audio signals carry structured temporal-frequency information. Motion signals reflect lower-dimensional physical dynamics. Neuromorphic sensors produce sparse asynchronous events. Wireless sensing, including WiFi channel state information and millimeter-wave radar, captures motion-related information through time-varying channel dynamics (14,15). These differences imply that SNN advantages may be modality-dependent rather than universal. Yet current evidence remains fragmented, because most prior studies focus on individual neuron models, encoding strategies, architectures, or datasets under inconsistent training and deployment protocols (16-18).

Here, we address this gap by constructing a modality-aware benchmark for SNN-based edge sensing. The benchmark covers five sensing modalities, 24 datasets, seven physical edge devices, and four neuromorphic simulation platforms. It systematically evaluates spike encoding strategies, neuron models, and network architectures under unified training and deployment protocols, and reports accuracy, inference latency, energy consumption, and memory footprint across 4,662 benchmark and deployment evaluations. This design allows us to move beyond asking whether SNNs are generally better than ANNs, and instead ask when and under what sensing conditions SNNs provide practical benefits.

Our results show that SNN benefits are selective and strongly modality-dependent. Across most sensing modalities, SNNs achieve performance broadly comparable to ANN baselines rather than uniformly surpassing them. Wireless sensing, however, emerges as a consistently favorable domain. Motivated by this observation, we further analyze wireless workloads using spectral and feature-space tools. These analyses suggest that the wireless advantage is associated with an alignment between the signal structure of wireless channel dynamics and the low-pass filtering behavior induced by spiking neuronal dynamics. This mechanism does not imply that all sensing modalities benefit from the same effect; rather, it illustrates how modality-specific signal dynamics can determine whether SNN computation is advantageous.

We further show that the consideration of SNN deployment and application should be multidimensional. Energy improvements do not automatically imply lower latency or memory usage, and algorithmic gains do not necessarily translate into system-level advantages on all hardware platforms. By jointly analyzing accuracy, latency, energy, and memory, our benchmark reveals that SNN deployment should be evaluated as a trade-off across modality, model design, encoding strategy, and hardware substrate.

Finally, we release Soul–NeuSim, an open-source benchmarking and deployment framework that integrates multimodal data processing, SNN model construction, edge deployment profiling, and neuromorphic simulation. Soul–NeuSim provides a reproducible infrastructure for comparing SNNs across sensing tasks and hardware settings, and supports future algorithm-software-hardware co-design. Together, these findings show that SNNs benefit edge sensing selectively rather than universally, with their advantages shaped by modality-specific signal dynamics and deployment constraints.

## RESULTS

### A modality-aware benchmark for SNN edge sensing

To determine when spiking neural networks (SNNs) benefit edge sensing, we construct a benchmark and deployment framework for comparing SNNs across heterogeneous sensing workloads and hardware settings. As shown in Fig. 1, the framework organizes SNN evaluation around three major design dimensions: spike encoding, neuronal dynamics, and network topology. First, we evaluate representative spike encoding strategies that transform continuous-valued inputs into spike trains, enabling systematic comparison of how encoding affects the preservation of task-relevant information. Second, we compare representative variants of leaky integrate-and-fire (LIF) neuron models (19), which are widely adopted in on-device SNNs because they offer a practical balance between simplicity and accuracy, to quantify how neuronal dynamics affect performance and efficiency. Third, we assess a diverse set of network topologies, ranging from convolutional backbones to transformer-inspired spiking models, to examine how connectivity patterns and temporal modeling capacity influence practical deployment (20-23).

We applied this framework across five sensing modalities and evaluated both model-level accuracy and deployment-level behavior under unified protocols. This design allows us to compare SNNs and artificial neural networks (ANNs) not only by average accuracy but also by how their relative benefits change with modality, model design, and hardware constraints.

The resulting benchmark provides the basis for three analyses. We first examine how SNN design choices behave across sensing modalities. We then focus on wireless sensing, the modality in which SNNs show the most consistent advantage, and analyze its signal structure in relation to spiking dynamics. Finally, we evaluate whether algorithmic benefits translate into deployment gains across edge and neuromorphic devices.

### SNN design choices exhibit strong modality dependence

As shown in Fig. 2a and Fig. S1, we first assess how spike encoding affects SNN performance across sensing modalities. Direct encoding achieved the highest relative accuracy across modalities and was therefore used as an upper-bound reference, although it is not spike-compatible and is not considered a deployable SNN input format (24). Among practical spike-compatible encodings, performance varied substantially by modality. Motion sensing maintained relatively high accuracy across most encoding schemes, suggesting that its task-relevant information is more robust to spike conversion. Wireless sensing, in contrast, was more sensitive to encoding choice. Performance dropped under Burst and temporal-switch (TS) coding (25, 26), whereas time-to-first-spike (TTFS) and phase coding substantially outperformed rate coding (27,28). These results indicate that the information lost or preserved during spike conversion is modality-dependent rather than uniform across sensing workloads. Detailed encoding definitions and per-dataset results are provided in Supplementary Materials Sections 1 and 2.1.

As shown in Fig. 2b and Fig. S2, we next compare spiking neuron models across datasets within each modality. No single neuron model dominated across all modalities. In vision tasks, performance differences across neuron variants are relatively small (e.g., about 3.45% on CIFAR10 and nearly saturated on MNIST). In contrast, wireless sensing exhibits the largest variation, with TLIF improving over LIF by up to 4.79% on BullyDetect (29,30). Motion sensing shows stable rankings with marginal gaps (within 0.45%). Acoustic and neuromorphic tasks fall in between: certain variants (e.g., RPLIF ) consistently remain in the upper tier, while less suitable designs (e.g., TLIF or PSN) can lead to noticeable degradation, up to 5.99% on UrbanSound8K (31-33). These results show that neuronal dynamics interact with modality-specific signal properties and cannot be selected independently of the sensing task. Detailed per-dataset results are provided in Supplementary Materials Section 2.2.

As shown in Fig. 2c and Fig. S3, we then evaluate representative SNN architectures across modalities. Transformer-style SNNs, including Spikformer (34) and QKFormer (35), achieved the strongest overall performance and showed more consistent behavior across the evaluated modalities than conventional spiking models. However, architecture alone did not determine whether SNNs outperformed ANNs. Detailed per-dataset architecture results are provided in Supplementary Materials Section 2.3. As shown in Fig. 2d, when SNNs and ANNs were compared using the same LeNet backbone, vision, acoustic, motion, and neuromorphic tasks showed broadly comparable accuracy between the two paradigms. The largest and most consistent gap appeared in wireless sensing, where SNNs outperformed ANN counterparts under the matched-backbone comparison. These results reveal that SNN advantages are selective and modality-dependent, rather than universal across edge sensing workloads.

**Wireless sensing emerges as a favorable domain for SNNs**

Because the cross-modal benchmark identifies wireless sensing as the modality in which SNNs most consistently outperform ANNs, we next extend the comparison to a broader set of wireless datasets. Table 1 summarizes the performance of ANN and SNN architectures across these datasets. Across the expanded wireless benchmark, the results systematically show the accuracy gap observed between conventional ANNs and SNNs, demonstrating both superior performance and markedly improved parameter efficiency for SNNs. For the fine-grained activity datasets (Fi-HAR and UT-HAR) (36), SNNs reach near-ceiling performance (e.g., SEW-ResNet18 (37) and SpikingResFormer (23) achieving 100.0% and 99.73%, respectively), while matching or exceeding the strongest ANN baselines (e.g., BiLSTM (38) at 99.69% and ResNet18 at 98.11%). Several of these gains are achieved with substantially smaller models. For example, relative to larger ANN baselines such as ResNet50, Spikformer uses only 0.45M parameters, compared with 23.55M, corresponding to an approximate 52× reduction.

On more challenging wireless datasets, the advantage of SNNs becomes more pronounced and consistent. On Widar, the best SNN (SpikingResFormer, $81.42_{\pm0.32}$%) surpasses the strongest ANN (ResNet18, $71.70_{\pm0.33}$%) by 9.72%. On BullyDetect, QKFormer achieves $85.58_{\pm0.39}$%, exceeding the best ANN result (ResNet50, $73.48_{\pm1.72}$%) by 12.10 points. The most substantial improvements are observed in hand gesture and mmWave tasks: Spikformer attains $95.12_{\pm0.42}$% on AOPHand compared to the ANN best of $72.54_{\pm0.80}$% (a 22.58-point margin), while QKFormer achieves $80.00_{\pm2.50}$% on MMActivity, outperforming ViT ($65.83_{\pm5.77}$%) by 14.17 points.

These improvements are not driven by brute-force scaling. Several top-performing SNNs achieve the highest accuracy in this benchmark while maintaining compact parameter counts (e.g., QKFormer with 1.62M and Spikformer with 0.45M parameters). In contrast, extremely lightweight recurrent ANNs (e.g., RNNs with 0.03M parameters) show substantial performance degradation, highlighting the trade-off between model size and accuracy. Collectively, these results place SNNs on a more favorable accuracy-complexity frontier for wireless sensing, with transformer-style spiking architectures showing the strongest performance on the more demanding wireless tasks.

**Spectral and feature-space analyses explain the wireless advantage**

We next investigate why wireless sensing is a favorable domain for SNNs. We first analyse their processing dynamics in the frequency domain. Using a synthesized multi-tone signal $x(t) = \frac{1}{3}\,[\sin(2\pi \cdot 100t) + \sin(2\pi \cdot 200t) + \sin(2\pi \cdot 300t)]$ (Fig.3a) and its Fourier Transformed spectra (Fig.3b), we show that the LIF neuron attenuates high-frequency components and behaves as a temporal low-pass filter (39). In contrast, the standard ReLU activation in ANNs preserves more high-frequency components. Moreover, results from real-world data further support this low-pass behavior (Fig.3c-e). As shown in Fig.3d, the frequency-domain spectrum of the same feature map after LIF and ReLU processing differs markedly. The GradCAM (40) visualizations in Fig.3e show that SNNs emphasize global low-frequency features while suppressing high-frequency components.

These frequency-domain characteristics are consistent with the spectral structure of wireless sensing signals. As shown in the feature-time, feature-Doppler, and micro-Doppler spectrograms for the ARIL dataset (Fig.3f), task-relevant kinematic and environmental information is primarily embedded in the energy envelopes and Doppler shifts (processing details are provided in Supplementary Materials Section 3). Consequently, the discriminative spectra ($DI_{norm}$) for these modalities (Fig.3g) reveal that the most informative features are predominantly concentrated in the lower frequency bands, whereas high-frequency bands are largely dominated by environmental scattering noise.

These observations support the interpretation that the superior performance of SNNs on wireless datasets is associated with a natural "spectral alignment" between the neuronal dynamics and the discriminative spectrum of the data. As shown in Fig. 3h, the decay constant $\beta$ strongly affects the Direct Current-normalized power profile of the neuron on the wireless dataset (41); corresponding analyses for the mmWave dataset AOPHand are provided in Supplementary Materials Section 5. Increasing neuron's $\beta$ suppresses all non-DC frequency components monotonically and narrows the effective passband. By combining this characteristic with $DI_{norm}$, we can quantify "spectral alignment" using the Frequency-Matching Score (FMS), whose definition is detailed in Supplementary Materials Section 4. As shown in Fig. 3i, the FMS is consistently higher in low-frequency bins across all $\beta$ settings, which aligns with the low-frequency dominance indicated by $DI_{norm}$. In particular, unlike traditional low-pass filters that truncate energy beyond the $-3dB$ cutoff for high-frequency components, the SNN could effectively exploit those discriminative features for correct classification.

Fig. 3j presents some wireless sensing samples correctly classified by both networks (top row) versus those recognized only by the SNN (bottom row). The ANN correctly recognizes samples with clear trends in the top-10 channels, but struggles when the signals fluctuate heavily. In these noisy samples, where low-frequency gesture signals are mixed with high-frequency noise, the SNN could filter out the high-frequency noise to classify based on clean motion data. In contrast, the ANN keeps high-frequency noise as part of informative features, leading to feature contamination and misclassification. T-SNE visualizations (42) can further illustrate such feature contamination. As shown in Fig. 3k, features extracted by the SNN form more separable clusters compared to the ANN on the same dataset, achieving a silhouette score (43) of 0.60. The latent representations generated by the ANN for the wireless dataset exhibit extensive overlap, resulting in highly ambiguous class boundaries and yielding a silhouette score of only 0.39 (approximately 0.20 lower than that of SNN).

We further compare the representational fidelity of the two networks using sample reconstruction. The results in Fig. 3l show that the ANN reconstructs the sample with few useful features (in yellow) while mostly producing noise (in black). By contrast, the SNN, acting as a low-pass temporal filter, suppresses noise during reconstruction and produces the sample with greater fidelity. These observations suggest that the effectiveness of SNNs in wireless sensing depends on how well spike encoding, neuronal dynamics, and architectural bias align with the signal's spectral structure.

**Deployment benefits are multi-dimensional and hardware dependent**

To determine whether model-level advantages translate into practical deployment benefits, we evaluated SNNs across physical edge devices and simulated neuromorphic platforms using inference latency, energy consumption, and peak memory footprint as system-level metrics.

Even when instantiated with the same architecture, ANNs and SNNs exhibit markedly different deployment profiles because of their distinct computational mechanisms. On conventional hardware, SNNs generally incurred higher latency because each input was processed over multiple sequential timesteps. This overhead was modality dependent. As shown in Fig. 4a, SNN latency was 3.07× that of ANNs for neuromorphic event-sensing workloads and 2.84× that of ANNs for wireless workloads, whereas the ratio decreased to 1.63× for motion sensing. The latency gap was smaller on GPU-accelerated devices than on CPU-based devices, suggesting that hardware parallelism can partly offset the sequential timestep overhead of SNN inference.

As shown in Fig. 4b, energy consumption showed a different pattern. Neuromorphic event-sensing workloads exhibited the largest energy reductions, reaching up to 29.58×, whereas vision workloads showed the smallest gains, with reductions as low as 1.27×. This pattern is consistent with the stronger temporal sparsity of neuromorphic data, whereas vision involves dense, frame-based inputs in which energy is dominated more by convolution and memory access than by sparse spike computation.

As shown in Fig. 4c, the memory footprint revealed an additional trade-off. SNNs showed higher mean memory usage across modalities because membrane states and intermediate temporal activations must be retained during inference. The largest mean increases were observed for neuromorphic event-sensing and wireless workloads. Neuromorphic exhibits the largest observed mean increase (65.56%; paired t-test, $p=0.10$), consistent with its longer effective temporal horizon. Wireless also shows substantial average overhead (41.91%; paired t-test, $p=0.13$). These results indicate that energy gains should not be interpreted as uniform deployment gains, because lower energy consumption may be accompanied by higher latency or memory demand. Detailed per-device comparisons of inference latency, energy and memory are provided in Figs. S4-S6 and Supplementary Materials Sections 6.1-6.3.

We further examined how SNN architectures respond to hardware heterogeneity. After normalizing latency across devices, convolutional backbones such as SpikingVGG and SEW-ResNet showed more compact latency distributions, whereas the transformer-style Spikformer exhibited broader variation, indicating greater sensitivity to hardware-specific execution behavior (Fig. 4d). Energy consumption also depended strongly on architecture and modality. As shown in Fig. 4e, lightweight backbones such as QKFormer occupied lower-energy regimes and consumed 64.79–91.83% less energy than larger transformer-style or deeper architectures, with the strongest advantage observed in wireless tasks. Memory usage was shaped by both device and modality, with wireless sensing imposing the highest average memory demand and motion sensing the lowest (Fig. 4f). Detailed architecture-level inference latency, energy, and memory results are provided in Figs. S7 and S8 and Supplementary Materials Sections 6.4-6.6.

We also evaluated SNN deployment using neuromorphic hardware simulation (Fig. 4g). Simulated latency varied across platforms because of differences in spike-routing architecture and neuron-core parallelism. Energy consumption was positively associated with latency, consistent with event-driven computation in which energy dissipation is tied to spike activity. Memory usage was more strongly determined by workload than by chip configuration, with wireless and acoustic sensing imposing larger memory demands than motion sensing. Overall, these results show that SNN deployment should be evaluated as a multi-objective trade-off across accuracy, latency, energy, and memory, rather than as a single energy-efficiency claim. Figs. S9 and S10, together with Supplementary Materials Sections 6.7-6.9, provide additional detailed analyses.

**Soul-NeuSim supports reproducible benchmarking and neuromorphic co-design**

Finally, as shown in Fig.5, we integrated the benchmark and deployment pipeline into Soul-NeuSim, an open-source framework for reproducible evaluation of SNNs in edge sensing systems. Soul provides the front-end benchmarking toolkit, including standardized data processing, spike encoding, model construction, training configuration, and deployment-oriented evaluation. It supports heterogeneous sensing inputs and logs configurations, random seeds, hyperparameters, and profiling outputs to reduce uncontrolled experimental variability.

NeuSim provides the back-end deployment and neuromorphic simulation stack. It translates trained SNNs into hardware-constrained representations, models neuromorphic cores and network-on-chip routers, and supports cycle-level estimation of inference latency, communication behavior, resource utilization, and energy consumption. The back end also includes compilation components for synapse extraction, partitioning, and physical-core mapping, enabling controlled exploration of workload–architecture interactions.

Together, Soul and NeuSim provide a unified workflow spanning algorithm development, deployment profiling, and neuromorphic hardware simulation. This framework allows future studies to add new datasets, neuron models, architectures, encoding schemes, and hardware configurations under a consistent evaluation protocol. By releasing Soul–NeuSim, this work provides a reproducible basis for modality-aware comparison of SNNs and for algorithm-software-hardware co-design in edge intelligence.

## DISCUSSION

This study provides a modality-aware benchmark for evaluating when spiking neural networks provide practical benefits for edge sensing. Across diverse sensing workloads, SNNs did not show a uniform advantage over ANNs. Instead, their benefits depended on the interaction among sensing modality, spike encoding, neuronal dynamics, network topology, and deployment substrate. This finding shifts the question from whether SNNs are generally better than ANNs to when their temporal and event-driven computation is well matched to the structure of the sensing task (44).

A central result is that wireless sensing emerged as the clearest favorable domain for SNNs. In most other modalities, SNNs achieved accuracy broadly comparable to ANN baselines, but the advantage was less consistent and more dependent on model configuration. Wireless workloads showed stronger sensitivity to spike encoding and neuronal dynamics, and several spiking architectures outperformed ANN counterparts by a substantial margin. This pattern suggests that SNN benefits are most likely to appear when the input modality contains temporal structure that can be effectively preserved by spike encoding and processed by neuronal dynamics.

The wireless analysis provides a mechanistic interpretation of this result. LIF-type spiking neurons behave as temporal dynamical filters, attenuating high-frequency components while preserving slower variations. Wireless sensing signals, including WiFi channel state information and millimeter-wave radar, often contain discriminative information in time-varying channel responses with modality-specific spectral characteristics (45,46). Our spectral analysis, feature-space visualization, and reconstruction results consistently suggest that the wireless advantage arises from an alignment between the spectral-temporal structure of the input and the filtering behavior of spiking dynamics. This interpretation does not imply that SNNs are intrinsically superior for all temporal signals. Rather, it indicates that SNNs can become particularly effective when encoding, neuron dynamics, and architectural bias jointly emphasize the frequency bands that carry task-relevant information.

This perspective also helps explain why SNN performance varied across design choices. Spike encoding determines which information is preserved when continuous-valued sensor data are converted into spike trains. Neuronal dynamics determine how temporal evidence is accumulated, filtered, and reset (47). Network topology determines how spatial, temporal, and cross-channel dependencies are represented. These components are often evaluated separately in prior studies, but our results show that their effects are modality-dependent and mutually coupled. For practical SNN design, selecting an encoding scheme, neuron model, or backbone architecture without considering the signal characteristics of the target modality may lead to misleading conclusions.

The deployment results further show that SNN advantages cannot be reduced to energy efficiency alone (48). On conventional edge devices, SNNs often incurred higher latency because inputs were processed over multiple timesteps. They also required additional memory to maintain membrane states and intermediate temporal activations. At the same time, SNNs achieved substantial energy reductions in temporally sparse workloads, especially neuromorphic event-sensing tasks. These results reveal a multi-objective trade-off in which an SNN may reduce energy consumption while increasing latency or memory footprint. Therefore, the optimization problem of SNN research should jointly consider accuracy, latency, energy, and memory rather than relying on accuracy or energy estimates alone.

Hardware heterogeneity further shaped these trade-offs. GPU-accelerated devices partly reduced the latency penalty of multi-timestep SNN inference, whereas CPU-based platforms exposed stronger sequential execution overhead (49,50). Different SNN backbones also responded differently to hardware variation: convolutional spiking models showed more stable latency distributions, whereas transformer-style spiking models were more sensitive to device-specific execution behavior. Simulated neuromorphic platforms provided an additional view of deployment behavior, showing that routing architecture, neuron-core parallelism, and workload characteristics jointly affect latency, energy, and memory (24, 51-53). These findings support a hardware-aware view of SNN design, in which model selection and hardware substrate should be considered together.

Soul−NeuSim was developed to support this type of evaluation. By integrating multimodal data processing, spike encoding, neuron-model construction, network benchmarking, physical edge-device profiling, and neuromorphic simulation, the framework provides a reproducible pipeline for comparing SNNs across both algorithmic and system-level dimensions. Its purpose is not only to rank existing SNN models, but also to provide an infrastructure for studying how sensing modality, model design, and deployment platform interact. This is important because future edge systems will increasingly combine heterogeneous sensors, constrained hardware, and workload-specific accuracy and efficiency requirements.

Several limitations and future directions remain. First, although the benchmark covers five representative sensing modalities and a broad set of datasets, it does not exhaust the full diversity of edge sensing tasks. Other domains, such as biomedical signals (54), tactile sensing (55), industrial monitoring (56), and multimodal fusion workloads (57), may exhibit different signal statistics and deployment trade-offs. Second, our mechanistic analysis is intentionally focused on wireless sensing, because this modality showed the most consistent SNN advantage in the benchmark. However, this does not imply that the same spectral framework is sufficient for all sensing modalities. Vision, acoustic, motion, and neuromorphic event data may require different theoretical descriptions. Third, our mechanistic interpretation focuses primarily on LIF-type neuronal dynamics, which are widely used in edge-oriented SNNs because of their simplicity and hardware compatibility. Although the benchmark compares multiple LIF-family variants, extending the same mechanistic analysis to more complex adaptive, recurrent, or biologically detailed neurons remains an important direction for future work. Fourth, deployment measurements on conventional edge devices and simulations of neuromorphic platforms provide complementary but incomplete views of real-world deployment. Future studies should include more measurements on physical neuromorphic hardware, runtime systems, and compiler stacks as they become more accessible. Finally, this study focuses primarily on supervised sensing tasks; continual, adaptive, and online learning scenarios may introduce additional constraints on memory, latency, and energy.

Overall, this work shows that SNNs benefit edge sensing selectively rather than universally. Their practical value depends on the alignment between spiking temporal dynamics and modality-specific signal structure, as well as on whether the resulting models can satisfy deployment constraints. This modality-aware view suggests that progress in SNN edge intelligence requires co-design across data representation, spike encoding, neuronal dynamics, network topology, and hardware execution. By providing a unified workflow for benchmarking, edge-device evaluation, and neuromorphic simulation, Soul-NeuSim offers a reproducible basis for studying such co-design in emerging edge and neuromorphic computing systems.

## MATERIALS AND METHODS

### Experimental design

We designed Soul-NeuSim to benchmark when spiking neural networks (SNNs) provide practical advantages for edge sensing. The study compared SNNs with artificial neural network (ANN) baselines across heterogeneous sensing modalities, model designs, and deployment substrates under controlled training and evaluation protocols. The experimental design contained four components. First, we evaluated three major SNN design dimensions: spike encoding, neuronal dynamics, and network topology. Second, we compared SNNs and ANNs across representative sensing modalities to identify modality-dependent performance trends. Third, because wireless sensing showed the most consistent SNN advantage, we performed additional wireless experiments and frequency-domain analyses to examine the relationship between wireless signal structure and spiking dynamics. Fourth, we profiled deployment behavior on physical edge devices and simulated neuromorphic platforms using accuracy, inference latency, energy consumption, and memory footprint as system-level metrics.

The main benchmark of this study included five sensing modalities: vision, acoustic, motion, wireless, and neuromorphic event sensing. For each modality, four representative datasets were used. Additional four wireless datasets were included in the expanded wireless analysis to evaluate whether the observed SNN advantage generalized across WiFi channel-state-information and millimeter-wave sensing workloads. Unless otherwise specified, all models within a matched comparison used the same data split, preprocessing pipeline, training protocol, and deployment configuration. Experiments were repeated over three independent runs with fixed random seeds, and results are reported as mean ± standard deviation. To support reproducibility, a centralized workflow monitor in Soul-NeuSim records configuration files, random seeds, hyperparameters, device settings, and profiling outputs for each experiment.

**Datasets and preprocessing**

The benchmark covered conventional sensing modalities and neuromorphic event-based sensing. Vision datasets were used to represent frame-based spatial perception. Acoustic datasets were used to represent non-stationary time-frequency signals. Motion datasets were used to represent multivariate inertial time series. Wireless datasets were used to represent contact-free sensing based on channel-state or radar-derived measurements. Neuromorphic datasets were used to represent sparse event streams generated by sensors such as dynamic vision sensors and artificial cochleae.

For datasets with official training and test splits, we followed the original split protocols (58-60). Vision inputs were kept at their original resolution unless established preprocessing conventions required resizing. For image datasets, standard augmentation such as random cropping and horizontal flipping was applied during training, followed by normalization. Acoustic signals were converted into mel-frequency spectrograms and resized to 128×128. Motion-sensing samples were constructed as multivariate time-series windows using a window length of 128 and a stride of 64. For wireless sensing, amplitude information was extracted from the raw complex channel state information when applicable. Neuromorphic event data were converted into frame-based representations at discrete time steps, following established event-processing protocols.

Except for neuromorphic event data, inputs from the other modalities were converted into temporally structured inputs before SNN processing. A default simulation length of four timesteps was used for non-neuromorphic workloads unless otherwise specified. Neuromorphic workloads used 10 timesteps to preserve their native event-driven temporal structure (61). Detailed preprocessing procedures, dataset splits, and temporal settings are provided in Supplementary Materials Section 7.

**Spike-encoding strategies**

We evaluated representative spike-encoding strategies to examine how continuous-valued sensor inputs are transformed into temporal spike trains. The encoding set included rate coding, time-to-first-spike coding, burst coding, phase coding, and temporal-switch coding. These methods generate spike-wise binary event streams over a temporal window and therefore represent deployable spike-compatible input strategies.

Direct coding was also evaluated as an upper-bound reference. In direct coding, the normalized input is replicated over multiple timesteps and injected as a constant analog drive. This strategy preserves input precision and often achieves strong software benchmark performance, but it relies on real-valued synaptic input rather than discrete spike events. Therefore, direct coding was not treated as a spike-compatible input format for spike-only neuromorphic hardware (24,51).

For each dataset, the timestep length and encoding parameters were selected according to the modality and input format. The same encoding configuration was used across models within a given benchmark setting to ensure comparability. Detailed encoding definitions and parameter settings are provided in Supplementary Materials Section 8.

### Neuron models and surrogate-gradient training

To evaluate the role of neuronal dynamics, Soul-NeuSim decouples neuron models from network topology through a modular neuron factory. The framework supports 13 spiking neuron models and 13 surrogate-gradient formulations. The evaluated neuron models include representative LIF-family and related spiking neuron variants that differ in temporal integration, leakage, thresholding, reset behavior, and adaptive dynamics. These models were used to quantify how neuron-level temporal dynamics affect accuracy and deployment behavior across sensing modalities. A detailed description of the implemented neuron models is provided in Supplementary Materials Sections 9 and 14.

SNNs were trained using surrogate-gradient backpropagation (47). The benchmark framework supports multiple surrogate-gradient formulations, and the arctangent surrogate function was used as the default in the benchmark unless otherwise specified. By holding the network topology and training settings fixed while changing neuron models, the benchmark isolates the effect of neuronal dynamics from architecture-level confounders.

### Network architectures and ANN baselines

The benchmark framework provides 12 SNN backbones and hybrid models under a unified interface. We evaluated representative SNN backbones spanning convolutional, residual, transformer-based, and hybrid attention-based design paradigms. The evaluated SNNs included SpikingVGG, SEW-ResNet, MS-ResNet, Spikformer, Meta-Spikeformer, QKFormer, and SpikingResFormer. These architectures cover conventional convolutional feature extraction, residual spiking computation, membrane-level shortcuts, spiking self-attention, and hybrid residual-attention designs.

For the ANN-SNN comparison, LeNet was used as a common lightweight baseline architecture. This choice reduces architecture-level bias and allows direct comparison between ANN activations and SNN neuronal dynamics under the same architecture. For broader topology comparisons, SNN architectures were adapted to non-vision sensing inputs while preserving their core architectural design. Model capacity was controlled either by matching parameter counts or by aligning computational complexity when exact parameter matching was not possible. A detailed description is provided in Supplementary Materials Section 10.

### Training protocol

All models were trained under consistent optimization settings within each dataset unless otherwise specified. We used the Adam optimizer (62) with a cosine annealing learning-rate scheduler (63) and an initial learning rate of $1\times10^{-3}$ . Training was performed on NVIDIA RTX 4090 GPUs with a batch size of 16. Each model was trained for 150 epochs to obtain stable test accuracy. Detailed configurations, including the optimizer, learning rate scheduler, training epochs, and other settings, are provided in Supplementary Materials Section 11.

### Evaluation metrics

All sensing tasks were formulated as classification tasks. Model correctness was evaluated using top-1 classification accuracy. Deployment behavior was evaluated using inference latency, energy consumption, and peak memory footprint.

Inference latency was defined as the elapsed wall-clock time required to process a single input sample and produce the final prediction. Warm-up runs were performed before timing to reduce one-time overhead from kernel loading, memory allocation, and runtime initialization.

Timing measurements were conducted under identical batch-size and deployment settings. In addition to mean latency, P90 latency was used to characterize runtime variability.

Energy consumption was evaluated using the corresponding profiling or simulation procedure for each deployment target. Peak memory footprint was defined as the maximum runtime memory usage during inference, including model parameters, intermediate activations, and SNN temporal states. For operation-level comparison, ANN complexity was measured using FLOPs computed from layer-wise dense operations, whereas SNN complexity was measured using synaptic operations, which account for timestep count and presynaptic firing activity (64,65). The first SNN layer was treated as dense under direct input encoding, while subsequent layers were measured as spike-driven synaptic operations. Detailed metric definitions and computation procedures are provided in Supplementary Materials Section 12.

**Physical edge-device deployment**

After training, models were converted into device-compatible formats through a standardized deployment interface based on NCNN (66). This study evaluated seven physical edge devices, including Raspberry Pi 4B, Jetson Nano, Jetson NX, Jetson AGX, Redmi K80, Pixel 6, and Huawei Mate 40. These devices cover CPU-, GPU-, and NPU-based edge execution environments. Detailed descriptions for physical edge-device deployment are provided in Supplementary Materials Section 13.

For each deployment setting, the same trained model, input representation, and preprocessing pipeline were used whenever supported by the deployment backend. Model conversion, execution scheduling, and metric collection were separated to reduce device-specific optimization bias. Latency, energy consumption, and peak memory footprint were recorded under identical batch-size and numerical-precision settings within each device. Physical-device profiling was used to quantify how ANN and SNN deployment behavior changes with sensing modality, model architecture, and hardware substrate.

**Wireless spectral analysis**

To investigate why wireless sensing is a favorable domain for SNNs, we analyzed the frequency-domain structure of wireless sensing signals and compared it with the temporal filtering behavior of spiking neuronal dynamics. For wireless inputs with explicit temporal structure, we extracted temporal proxy signals by reducing non-temporal dimensions through mean pooling. The resulting sequences were transformed into the frequency domain using discrete Fourier transforms (67).

For millimeter-wave sensing data, we computed feature-time energy maps, feature-Doppler energy maps, and micro-Doppler spectrograms. Feature-time energy maps were computed from the instantaneous power of each feature channel. Feature-Doppler maps were obtained by applying a discrete Fourier transform along the time axis. Micro-Doppler spectrograms were computed using short-time Fourier transforms on selected high-variance temporal channels.

To quantify task-relevant frequency structure, we computed a discriminative spectrum based on the Fisher criterion (68). For each frequency bin, between-class and within-class scatter were estimated from the training set, and their ratio was used to define a discriminative index. The discriminative index was normalized over the one-sided frequency grid to form a probability mass function over frequency. This procedure estimates where class-discriminative information is concentrated in the frequency domain without relying on a specific neural architecture. Detailed wireless spectral-analysis procedures are provided in Supplementary Materials Sections 3 and 5.

**Frequency-matching score**

We used the frequency-matching score to quantify the alignment between wireless discriminative spectra and the frequency response induced by neuron dynamics (69). The discrete-time LIF neuron can be interpreted as a first-order infinite-impulse-response low-pass filter. Its membrane decay factor determines the strength of frequency attenuation and the effective passband of the neuron.

For each candidate membrane decay factor, we computed a DC-normalized power response of the neuron dynamics on the one-sided frequency grid. The frequency-matching score was then computed as the weighted fraction of discriminative spectral content retained by this normalized response. Higher scores indicate stronger overlap between the frequency bands emphasized by spiking dynamics and those carrying task-relevant information in the wireless input. We further used a maximum-deviation rule on the frequency-matching curve to define a reference boundary separating under-filtering, stable filtering, and over-low-pass regimes. The definition and full derivation of the frequency-matching score are provided in Supplementary Materials Section 4.

**Feature-space visualization and reconstruction analysis**

To compare ANN and SNN representations on wireless sensing tasks, we extracted latent features from trained models and visualized their distributions using t-distributed stochastic neighbor embedding. Feature separability was quantified using silhouette scores computed on the extracted latent representations. These analyses were used to assess whether SNN representations produced more separable class clusters than ANNs under the same wireless sensing task.

We also compared representative reconstruction outputs from ANN and SNN processing pipelines. The reconstruction analysis was used to qualitatively examine whether model representations retained noise-dominated components or preserved cleaner task-relevant structures. Together with the discriminative-spectrum and frequency-matching analyses, these representation-level analyses were used to interpret the wireless advantage of SNNs. Additional visualization and reconstruction analyses are provided in Supplementary Materials Section 5.

**Neuromorphic simulation with NeuSim**

NeuSim provides the back-end simulation stack for neuromorphic deployment analysis. It models contemporary neuromorphic processors as arrays of neuromorphic cores connected by a two-dimensional mesh network-on-chip (24, 51-53). SNN workloads are partitioned across cores for parallel execution (70). Hardware parameters, including neuron capacity, synapse memory size, buffer depth, chip dimensions, and routing policy, are configurable to support architectural exploration. Hardware parameter settings are summarized in Supplementary Materials Section 13.

NeuSim uses two main abstractions: neuromorphic cores and network-on-chip routers. Within each core, neuron units update neuronal states and maintain membrane potentials, synapse memory stores connectivity information and post-synaptic addresses, and the network interface encapsulates spikes into packets. Router modules model packet traversal across the interconnect, including routing, virtual-channel allocation, switch arbitration, and crossbar traversal. XY routing was used by default unless otherwise specified (71).

Core and router modules operate under cycle-level simulation to estimate computation and communication delays. The simulator records end-to-end inference latency, core-level resource utilization, communication traffic, routing congestion, packet-level transmission delay, and energy consumption. Energy estimates are obtained from configurable power models parameterized using empirical power data from representative physical hardware devices. NeuSim supports multi-threaded execution, with synchronization at every clock cycle to preserve temporal consistency, and exposes its C++ simulation engine to Python for integration with Soul.

**Compilation, partitioning, and mapping**

To deploy trained SNNs onto hardware-constrained neuromorphic architectures, NeuSim includes an automated compilation pipeline consisting of synapse extraction, partitioning, and physical-core mapping. The extraction stage transforms high-level neural network layers into neuron-synapse connectivity required by the hardware model. Rather than constructing a dense global adjacency matrix, the network is represented as a directed acyclic graph in which nodes correspond to neuron layers and edges correspond to operators. Linear layers are converted into fully connected bipartite connectivity patterns, whereas convolutional layers are transformed into structured local connectivity using unfolding-based extraction. Operator compositions are analytically fused to obtain final synaptic weights.

Partitioning groups neurons and synapses into logical cores subject to single-core neuron and synapse capacity constraints. Sequential partitioning was used by default. NeuSim also supports user-defined partitioning strategies, including a locality-preserving heuristic that maps neurons into a one-dimensional sequence using space-filling curves and then applies binary segmentation to produce capacity-valid clusters. Mapping then assigns logical cores to physical cores in the two-dimensional mesh topology. Default sequential placement and Hilbert space-filling curve (HSC)-based placement are both supported (72). These compiler interfaces allow customized partitioning and placement strategies to be integrated into the Soul-NeuSim workflow.

## Acknowledgments

**Funding:**
National Key Research and Development Program of China (2022YFB4500100)
National Key Research and Development Program of China (2025YFG0100700)
National Natural Science Foundation of China (62125206)
National Natural Science Foundation of China (62502443)

**Author contributions:**
Conceptualization: X.D., D.Y., S.D. and G.P.
Methodology: X.D., D.Y., C.L., Y.Z. and Z.C.
Investigation: X.D., D.Y., C.L., Y.Z., W.T., X.Z., L.J. and Z.L .
Visualization: D.Y., C.L., Y.Z., W.T., H.Z., W.Z. and S.J.
Supervision: X.D, J.F., S.D. and G.P.
Writing—original draft: X.D., D.Y, C.L, Y.Z, Z.C and W.T.
Writing—review & editing: X.D., H.F., J.F., H.T., X.Z., S.D. and G.P.

**Competing interests:**
All authors declare that they have no competing interests.

**Data and code availability:** All datasets analyzed in this study are freely available online resources. The code for the main results of this manuscript is already publicly available on GitHub at: https://github.com/yudi-mars/Soul-NeuSim.

## Figures and Tables

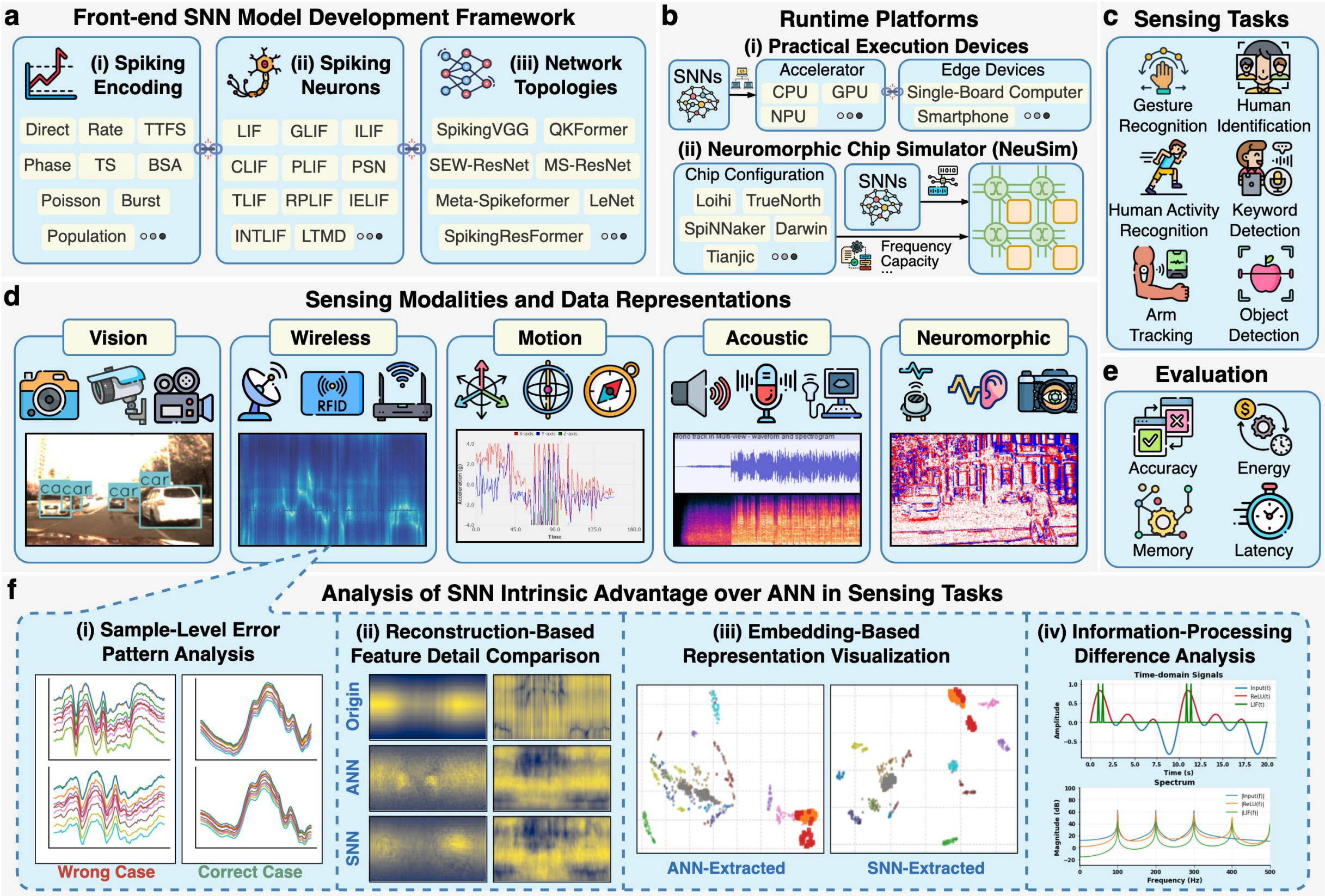

**Fig. 1. Overview of the modality-aware benchmarking and deployment framework.**
(a) A front-end model development framework for SNN benchmarking, supporting multiple spike encoding schemes, neuron models, and network topologies. (b) A back-end deployment platform for SNN benchmarking, including conventional edge devices with CPUs, GPUs, and NPUs, and neuromorphic chip simulation with NeuSim. (c) Representative edge-sensing tasks included in the benchmark. (d) Five sensing modalities and their data representations: vision, wireless, motion, acoustic, and neuromorphic event data. (e) Evaluation dimensions considered in the benchmarking pipeline include accuracy, energy consumption, memory footprint, and inference latency. (f) Wireless analysis workflow comparing ANN and SNN behavior through sample-level errors, reconstruction, feature embedding, and information-processing analysis.

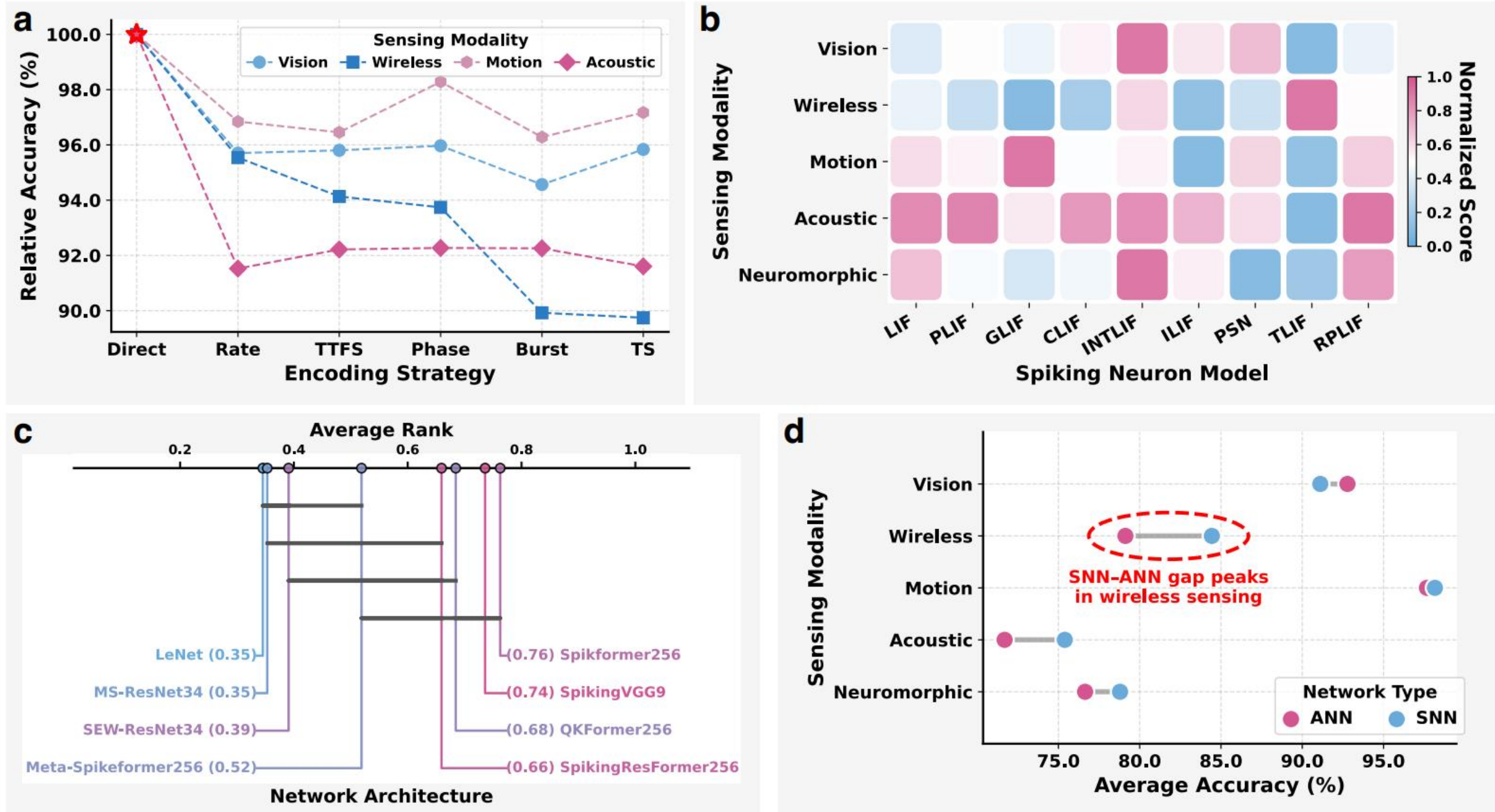

**Fig. 2. Performance landscape of spiking neural networks across sensing modalities, encoding strategies, neuron models, and architecture choices.** The five sensing modalities include vision, acoustic, motion, wireless, and neuromorphic sensing. Detailed per-dataset results and descriptions are provided in the Figs. S1-S3. (a) Relative accuracy of different spike encoding strategies across modalities. (b) Normalized performance of spiking neuron models across modalities. Colors indicate row-wise normalized accuracy to highlight the relative performance of neuron models within each modality. (c) Overall performance ranking of representative SNN architectures across all sensing modalities with a critical difference diagram (higher is better). (d) Average accuracy of SNNs and ANNs across modalities using the same LeNet backbone. The largest gap is observed in wireless sensing, where SNNs consistently outperform ANNs.

**Table 1. Evaluation of ANNs and SNNs with different architectures across wireless sensing datasets.** This table reports end-to-end test accuracy (%) and the corresponding number of model parameters in millions (#P., M) for commonly used ANN architectures and their SNN counterparts on the wireless sensing benchmark. For each dataset, the best-performing results are highlighted using **bold** and **bold**, while the second-best results are indicated with underline and underline.

| Dataset | Fi-HAR | | Widar | | UT-HAR | | Fi-HumanID | | BullyDetect | | ARIL | | AOPHand | | MMActvity | |
|---|---|---|---|---|---|---|---|---|---|---|---|---|---|---|---|---|
| **Method - ANN** | **Acc. (%)** | **#P. (M)** | **Acc. (%)** | **#P. (M)** | **Acc. (%)** | **#P. (M)** | **Acc. (%)** | **#P. (M)** | **Acc. (%)** | **#P. (M)** | **Acc. (%)** | **#P. (M)** | **Acc. (%)** | **#P. (M)** | **Acc. (%)** | **#P. (M)** |
| LeNet | 99.62$_{\pm 0.38}$ | 122.01 | 69.65$_{\pm 0.77}$ | 0.28 | 97.60$_{\pm 0.34}$ | 40.76 | 92.67$_{\pm 2.16}$ | 122.01 | 47.60$_{\pm 1.90}$ | 6.01 | 79.62$_{\pm 2.39}$ | 12.30 | 60.86$_{\pm 1.68}$ | 4.19 | 59.17$_{\pm 3.82}$ | 4.19 |
| ResNet18 | 95.31$_{\pm 0.22}$ | 11.18 | **71.70**$_{\pm 0.33}$ | 11.25 | **98.11**$_{\pm 0.06}$ | 11.18 | 96.42$_{\pm 0.01}$ | 11.19 | 73.08$_{\pm 2.34}$ | 11.18 | 89.21$_{\pm 1.25}$ | 11.17 | 72.47$_{\pm 1.30}$ | 11.17 | 61.67$_{\pm 6.29}$ | 11.17 |
| ResNet50 | 99.38$_{\pm 0.22}$ | 23.55 | 68.56$_{\pm 1.85}$ | 23.64 | 97.21$_{\pm 0.06}$ | 23.55 | 92.91$_{\pm 0.10}$ | 23.57 | **73.48**$_{\pm 1.72}$ | 21.29 | 87.89$_{\pm 4.17}$ | 21.28 | **72.54**$_{\pm 0.80}$ | 23.51 | 61.67$_{\pm 2.89}$ | 23.51 |
| ResNet101 | 95.31$_{\pm 1.09}$ | 42.57 | 68.71$_{\pm 1.94}$ | 42.66 | 94.99$_{\pm 0.06}$ | 42.57 | 88.40$_{\pm 0.32}$ | 42.59 | 73.05$_{\pm 0.43}$ | 42.59 | 85.37$_{\pm 7.41}$ | 41.35 | 69.70$_{\pm 0.72}$ | 42.51 | 64.17$_{\pm 1.44}$ | 42.50 |
| RNN | 84.64$_{\pm 0.95}$ | **0.03** | 47.05$_{\pm 0.66}$ | **0.03** | 83.53$_{\pm 6.36}$ | **0.01** | 89.30$_{\pm 0.64}$ | **0.03** | 24.12$_{\pm 0.56}$ | **0.01** | 25.90$_{\pm 0.72}$ | **0.01** | 26.40$_{\pm 9.25}$ | **0.03** | 48.33$_{\pm 1.44}$ | **0.03** |
| GRU | 97.66$_{\pm 1.74}$ | 0.08 | 62.50$_{\pm 0.54}$ | 0.09 | 94.18$_{\pm 1.88}$ | 0.03 | 98.96$_{\pm 1.25}$ | 0.08 | 47.81$_{\pm 0.24}$ | 0.03 | 54.08$_{\pm 5.85}$ | 0.02 | 67.52$_{\pm 5.73}$ | 0.08 | 47.50$_{\pm 2.50}$ | 0.08 |
| LSTM | 97.14$_{\pm 0.66}$ | 0.11 | 63.35$_{\pm 0.65}$ | 0.12 | 87.18$_{\pm 3.62}$ | 0.04 | 97.19$_{\pm 0.65}$ | 0.11 | 36.35$_{\pm 1.41}$ | 0.04 | 25.90$_{\pm 1.65}$ | 0.03 | 35.71$_{\pm 25.5}$ | 0.10 | 54.17$_{\pm 7.64}$ | 0.10 |
| BiLSTM | **99.69**$_{\pm 0.28}$ | 0.21 | 63.43$_{\pm 0.45}$ | 0.24 | 90.19$_{\pm 1.76}$ | 0.08 | **99.38**$_{\pm 1.29}$ | 0.21 | 38.01$_{\pm 2.34}$ | 0.08 | 37.53$_{\pm 4.22}$ | 0.06 | 38.68$_{\pm 13.0}$ | 0.20 | 45.83$_{\pm 3.82}$ | 0.20 |
| CNN-GRU | 93.75$_{\pm 0.22}$ | 0.06 | 63.19$_{\pm 0.41}$ | 0.09 | 96.72$_{\pm 0.11}$ | 1.43 | 87.48$_{\pm 0.87}$ | 0.06 | 71.06$_{\pm 1.12}$ | 0.14 | **91.49**$_{\pm 1.50}$ | 0.13 | 61.98$_{\pm 9.44}$ | 0.46 | 65.00$_{\pm 2.50}$ | 0.46 |
| ViT | 93.75$_{\pm 3.06}$ | 1.05 | 67.72$_{\pm 0.33}$ | 0.11 | 96.53$_{\pm 0.89}$ | 10.58 | 76.84$_{\pm 3.51}$ | 1.05 | 32.75$_{\pm 1.26}$ | 2.14 | 54.80$_{\pm 3.60}$ | 0.58 | 21.39$_{\pm 3.31}$ | 2.18 | **65.83**$_{\pm 5.77}$ | 2.18 |
| **Method - SNN** | **Acc. (%)** | **#P. (M)** | **Acc. (%)** | **#P. (M)** | **Acc. (%)** | **#P. (M)** | **Acc. (%)** | **#P. (M)** | **Acc. (%)** | **#P. (M)** | **Acc. (%)** | **#P. (M)** | **Acc. (%)** | **#P. (M)** | **Acc. (%)** | **#P. (M)** |
| LeNet | 99.75$_{\pm 0.44}$ | 122.01 | 76.78$_{\pm 0.28}$ | **0.28** | 98.66$_{\pm 0.21}$ | 40.76 | 96.15$_{\pm 1.02}$ | 122.01 | 56.22$_{\pm 0.30}$ | 6.01 | 85.73$_{\pm 0.91}$ | 12.30 | 87.72$_{\pm 2.72}$ | 4.19 | 75.83$_{\pm 3.82}$ | 4.19 |
| SEW-ResNet18 | **100.0**$_{\pm 0.00}$ | 11.18 | 81.30$_{\pm 0.26}$ | 11.25 | **99.73**$_{\pm 0.06}$ | 11.17 | 99.51$_{\pm 0.10}$ | 11.18 | 79.48$_{\pm 2.01}$ | 11.18 | 87.65$_{\pm 2.20}$ | 11.17 | 91.09$_{\pm 1.19}$ | 11.17 | 72.50$_{\pm 2.50}$ | 11.17 |
| SEW-ResNet50 | 98.86$_{\pm 0.38}$ | 23.52 | 80.11$_{\pm 0.30}$ | 23.61 | 99.57$_{\pm 0.06}$ | 23.52 | 97.99$_{\pm 1.29}$ | 23.54 | 77.59$_{\pm 0.56}$ | 23.52 | 86.21$_{\pm 1.78}$ | 23.51 | 90.69$_{\pm 1.39}$ | 23.51 | 74.17$_{\pm 5.20}$ | 23.51 |
| MS-ResNet18 | 99.87$_{\pm 0.22}$ | 11.18 | 79.53$_{\pm 0.15}$ | 11.25 | 99.57$_{\pm 0.06}$ | 11.18 | 99.33$_{\pm 0.28}$ | 11.19 | 75.85$_{\pm 0.36}$ | 11.18 | 88.85$_{\pm 0.95}$ | 11.18 | 91.22$_{\pm 2.10}$ | 11.18 | 70.00$_{\pm 11.5}$ | 11.18 |
| MS-ResNet50 | 96.97$_{\pm 3.01}$ | 23.52 | 78.68$_{\pm 0.44}$ | 23.61 | 99.40$_{\pm 0.30}$ | 23.52 | 98.90$_{\pm 0.32}$ | 23.54 | 73.78$_{\pm 0.46}$ | 23.52 | 89.33$_{\pm 1.70}$ | 23.51 | 90.56$_{\pm 0.41}$ | 23.51 | 74.17$_{\pm 3.82}$ | 23.51 |
| Spikformer | 99.73$_{\pm 0.06}$ | **0.45** | 78.26$_{\pm 0.45}$ | 0.45 | **99.73**$_{\pm 0.06}$ | **0.45** | 99.51$_{\pm 0.21}$ | **0.45** | 84.03$_{\pm 0.58}$ | **0.45** | **94.25**$_{\pm 1.56}$ | **0.45** | **95.12**$_{\pm 0.42}$ | 2.57 | 75.83$_{\pm 6.29}$ | 2.57 |
| Meta-Spikeformer | 98.10$_{\pm 2.01}$ | 3.86 | 81.29$_{\pm 0.06}$ | 3.89 | **99.73**$_{\pm 0.06}$ | 3.86 | 63.92$_{\pm 17.4}$ | 3.86 | 77.44$_{\pm 1.35}$ | 3.86 | 91.25$_{\pm 1.10}$ | 3.86 | 91.95$_{\pm 0.46}$ | 6.08 | 70.83$_{\pm 3.82}$ | 6.08 |
| QKFormer | 97.54$_{\pm 0.27}$ | 1.62 | 80.62$_{\pm 0.05}$ | 1.63 | 99.67$_{\pm 0.06}$ | 1.62 | **99.57**$_{\pm 0.10}$ | 1.62 | **85.58**$_{\pm 0.39}$ | 1.62 | 93.72$_{\pm 0.67}$ | 1.62 | 90.96$_{\pm 0.69}$ | **2.41** | **80.00**$_{\pm 2.50}$ | **2.41** |
| SpikingResFormer | **100.0**$_{\pm 0.00}$ | 10.81 | **81.42**$_{\pm 0.32}$ | 10.87 | 99.67$_{\pm 0.12}$ | 10.80 | **99.57**$_{\pm 0.10}$ | 10.81 | 85.39$_{\pm 0.58}$ | 10.81 | 84.65$_{\pm 0.42}$ | 10.80 | 91.35$_{\pm 0.75}$ | 17.31 | 72.50$_{\pm 2.50}$ | 17.31 |

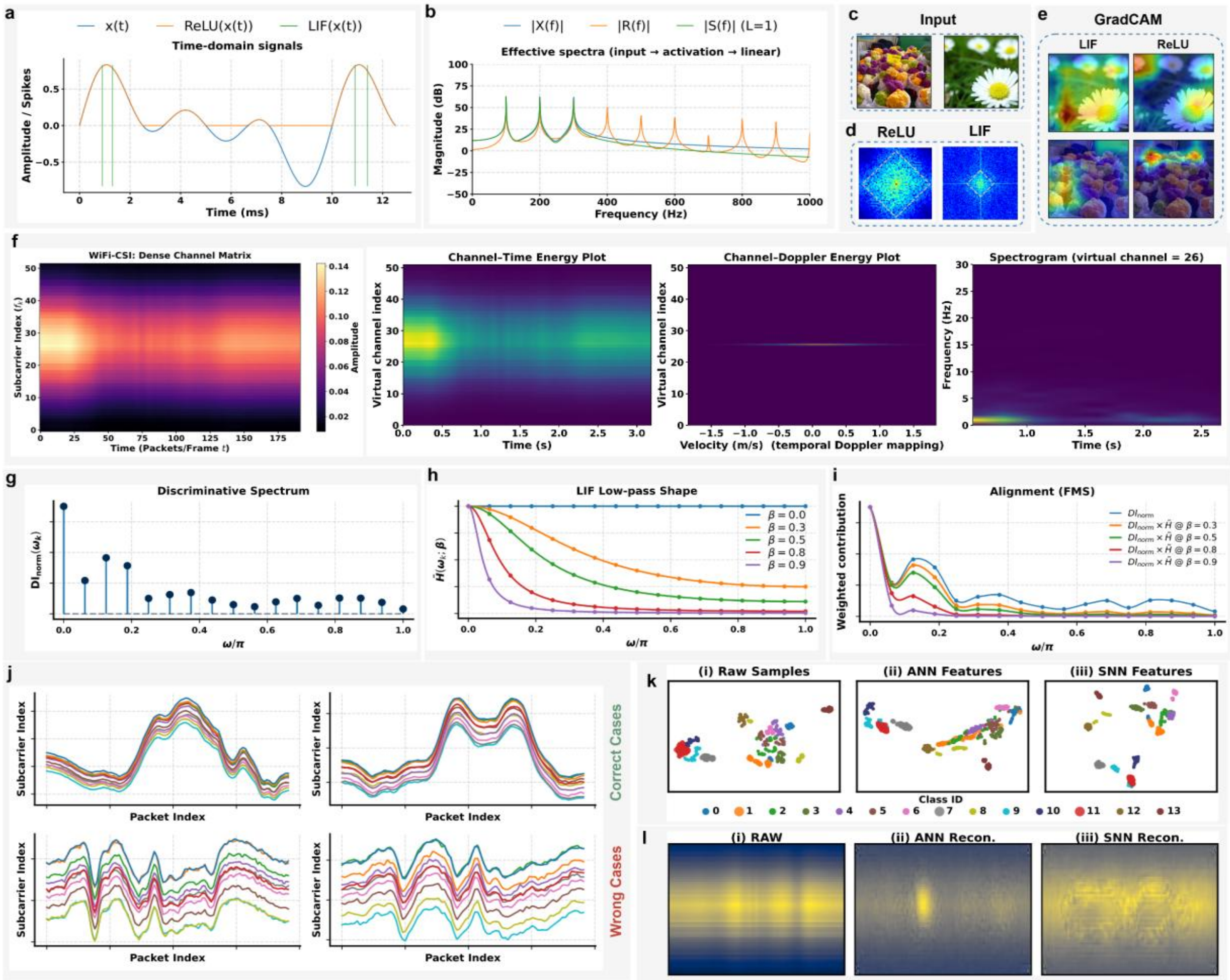

**Fig. 3. Spectral analysis of ANN and SNN dynamics and multi-level representation alignment of SNNs with wireless sensing.** (a) Time-domain visualization comparing the effects of ReLU and Leaky Integrate-and-Fire (LIF) activation functions on a synthesized multi-tone signal *x(t)*. (b) Corresponding frequency-domain magnitude spectra (dB) of the input x(t), its ReLU activation *R(x)*, and the extracted spike train S, computed via numerical FFT. (c)–(e) Vision sensing samples collected by RGB cameras (c) their spatial frequency spectra after the first encoding layer (d) and GradCAM class-discriminative visualizations (e) for ResNet18 and SEW-ResNet18. (f) Raw inputs, energy maps, and micro-Doppler spectrograms for the wireless ARIL. (g) Discriminative spectra for ARIL. (h)-(i) DC-normalized power templates (h) and Frequency- Matching Scores (FMS) (i) across different LIF membrane decay constants ($\beta$). (j) ARIL samples correctly classified by both networks (top row) and exclusively recognized by the SNN (*bottom row*). (k) t-SNE feature visualizations of the Fi-HumanID. (l) Encoder–decoder sample reconstructions for ARIL using shared structured ANN and SNN encoders and a common decoder backbone.

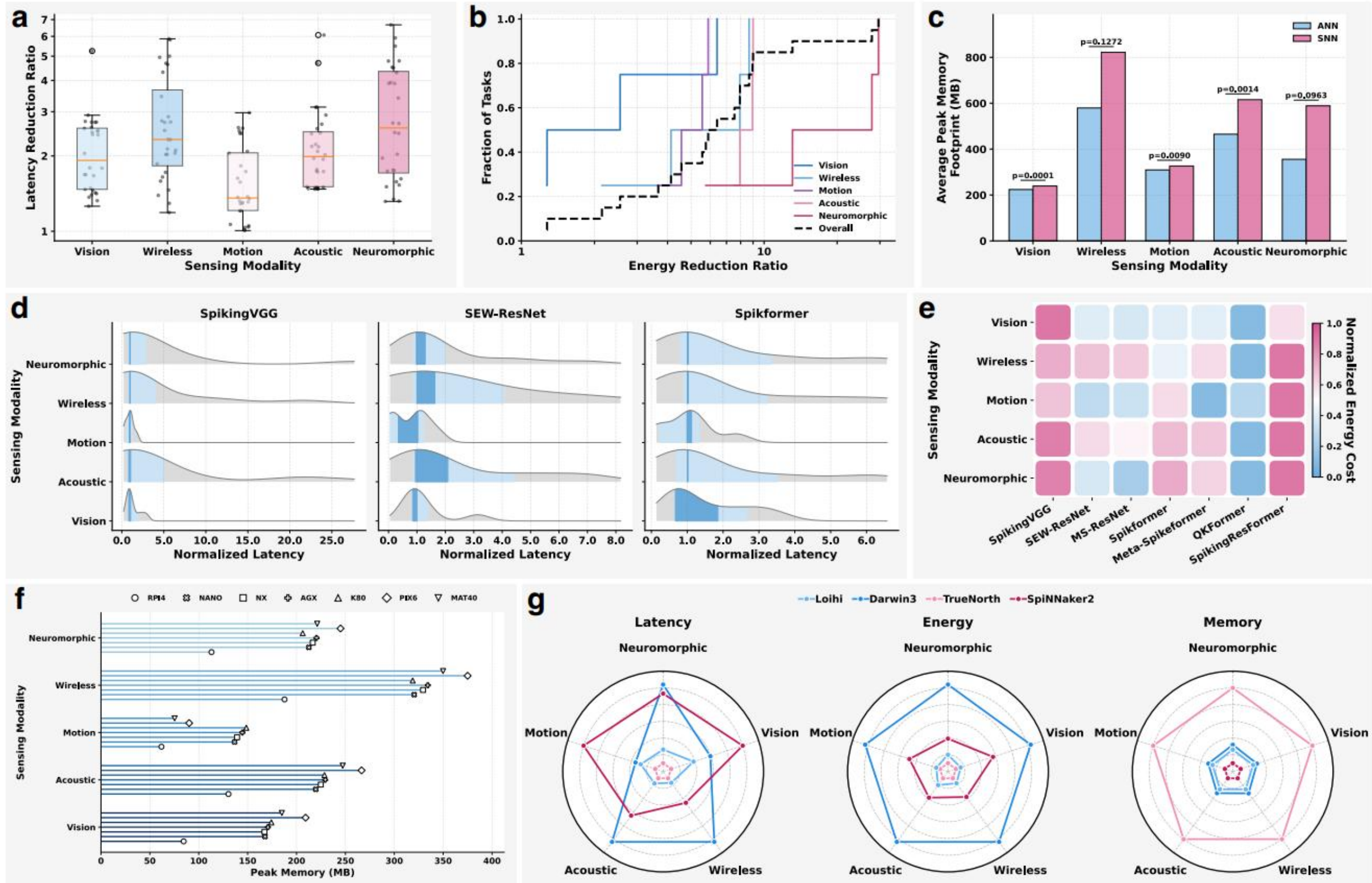

**Fig. 4 Deployment efficiency of spiking neural networks across sensing modalities and hardware devices.** (a) Distribution of latency reduction ratios (SNN/ANN) across five sensing modalities, showing that SNN inference is generally slower on conventional devices due to the recurrent computing paradigm, with variability across tasks. (b) Cumulative fraction of tasks achieving energy reduction (ANN/SNN) per modality, highlighting substantial energy savings, particularly for neuromorphic sensing; dashed line indicates overall distribution. (c) Average peak memory footprint (MB) of ANN vs. SNN, showing moderate memory overhead for SNNs, largest in wireless and neuromorphic tasks. (d) Normalized latency ranges for three representative SNN architectures (SpikingVGG, SEW-ResNet, Spikformer) across modalities. (e) Heatmap of normalized energy consumption for representative SNN architectures across modalities. (f) Average peak memory footprint across SNNs and tasks on edge devices. (g) Comparison of latency, energy, and memory across four neuromorphic simulation platforms (Loihi, Darwin3, TrueNorth, and SpiNNaker2) using MS-ResNet34 across modalities.

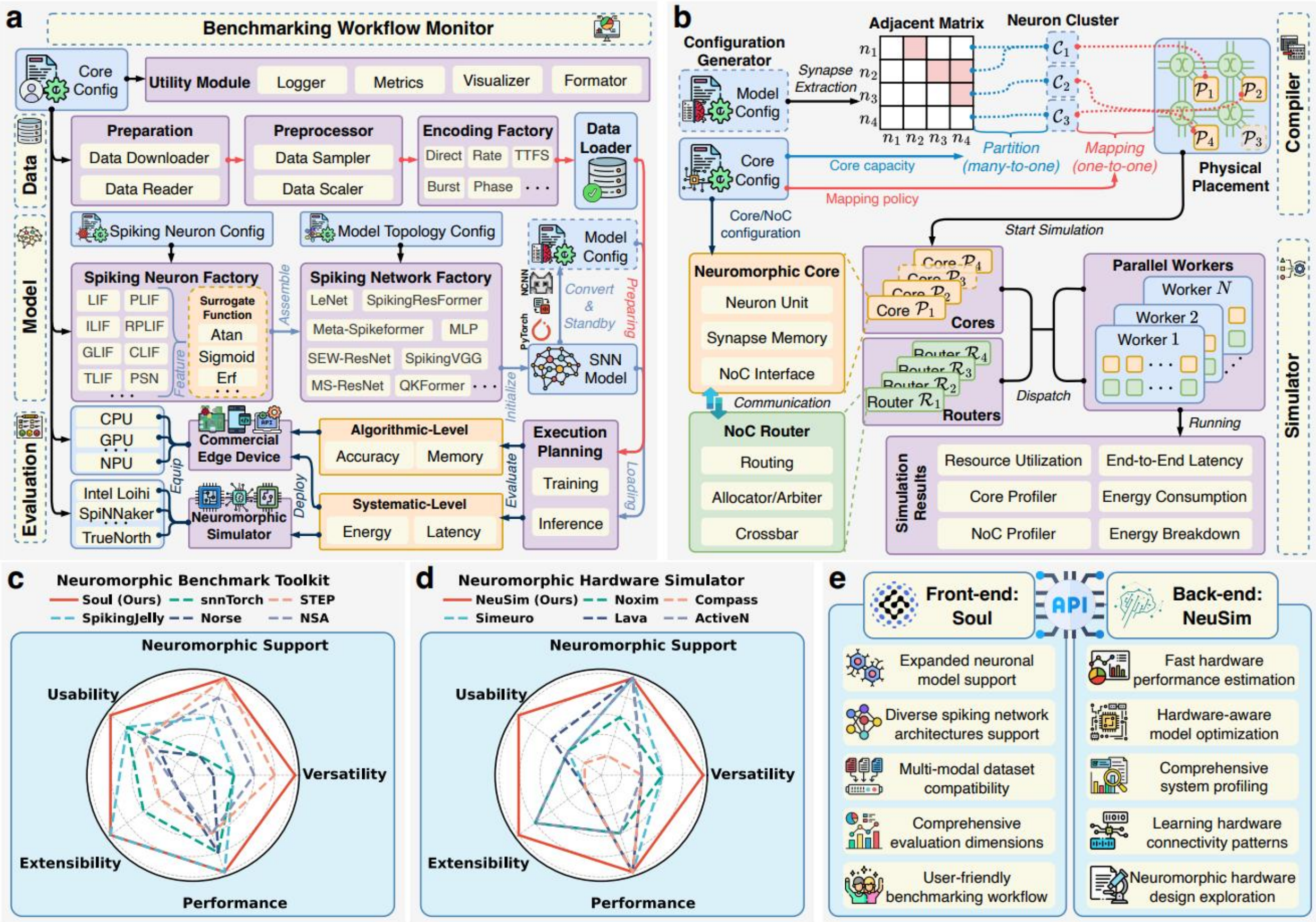

**Fig. 5. Soul–NeuSim framework for SNN benchmarking, deployment, and chip-level analysis.** (a) Overall architecture of the Soul front-end framework, illustrating a unified workflow that integrates multi-modal data processing, model initialization and preparation, and evaluation on heterogeneous edge devices. (b) Workflow for implementing SNNs in the NeuSim back-end simulator under different neuromorphic chip configurations. (c) Comparisons of Soul relative to existing frameworks. Soul provides extensibility for novel architectures and neurons, strong usability through clear documentation, broad support for neuromorphic models, versatility across multi-modal edge-sensing tasks, and training performance. (d) Comparisons of NeuSim relative to existing frameworks. NeuSim provides modular extensibility for custom hardware architectures, seamless usability through Python bindings and compiler integration, comprehensive support for diverse SNN models, versatility for mainstream neuromorphic chip configurations, and high-throughput performance enabled by a multi-threaded engine. (e) Soul–NeuSim co-designed software–hardware framework. Soul provides a frontend benchmarking framework for flexible SNN architectures, multimodal datasets, and unified evaluation, while NeuSim serves as the back end for fast hardware performance estimation, hardware-aware optimization, and design-space exploration, together enabling end-to-end algorithm–hardware co-design.

## Supplementary Materials

Supplementary Text
Figs. S1 to S10